\documentclass[sigconf]{acmart}

\usepackage{bbding}
\usepackage{algorithmic}
\usepackage{multirow}
\usepackage{amsfonts}
\usepackage{textcomp}
\usepackage{algorithm}
\usepackage{subcaption}
\usepackage{soul}
\usepackage{enumitem}
\soulregister{\cite}7 
\soulregister{\ref}7 
\soulregister{\mathbb}7

\usepackage{xcolor}

\AtBeginDocument{%
  }

\acmSubmissionID{129}

\acmISBN{978-1-4503-XXXX-X/18/06}

\copyrightyear{2024}
\acmYear{2024}
\setcopyright{acmlicensed}\acmConference[MMASIA '24]{ACM Multimedia Asia}{December 3--6, 2024}{Auckland, New Zealand}
\acmBooktitle{ACM Multimedia Asia (MMASIA '24), December 3--6, 2024, Auckland, New Zealand}
\acmDOI{10.1145/3696409.3700204}
\acmISBN{979-8-4007-1273-9/24/12}

\begin{document}

\title{SpikMamba: When SNN meets Mamba in Event-based Human Action Recognition}


\author{Jiaqi Chen}
\affiliation{%
  \institution{Northeastern University,}
  \city{Shenyang}
  \country{China}}
\email{2270691@stu.neu.edu.cn}

\author{Yan Yang}
\affiliation{%
  \institution{Australian National University,}
  \city{Canberra}
  \country{Australia}}
\email{Yan.Yang@anu.edu.au}

\author{Shizhuo Deng}
\affiliation{%
  \institution{Northeastern University,}
  \city{Shenyang}
  \country{China}}
\email{dengshizhuo@mail.neu.edu.cn}

\author{Da Teng}
\affiliation{%
  \institution{Northeastern University,}
  \city{Shenyang}
  \country{China}}
\email{13166672732@163.com}

\author{Liyuan Pan}
\authornote{Corresponding author.}
\affiliation{%
  \institution{Beijing Institute of Technology,}
  \city{Beijing}
  \country{China}}
\email{liyuan.pan@bit.edu.cn}

\renewcommand{\shortauthors}{Jiaqi Chen, Yan Yang, Shizhuo Deng, Da Teng, Liyuan Pan}

\begin{abstract}

Human action recognition (HAR) plays a key role in various applications such as video analysis, surveillance, autonomous driving, robotics, and healthcare. Most HAR algorithms are developed from RGB images, which capture detailed visual information. However, these algorithms raise concerns in privacy-sensitive environments due to the recording of identifiable features. Event cameras offer a promising solution by capturing scene brightness changes sparsely at the pixel level, without capturing full images. Moreover, event cameras have high dynamic ranges that can effectively handle scenarios with complex lighting conditions, such as low light or high contrast environments. However, using event cameras introduces challenges in modeling the spatially sparse and high temporal resolution event data for HAR. To address these issues, we propose the SpikMamba framework, which combines the energy efficiency of spiking neural networks and the long sequence modeling capability of Mamba to efficiently capture global features from spatially sparse and high a temporal resolution event data. Additionally, to improve the locality of modeling, a spiking window-based linear attention mechanism is used. Extensive experiments show that SpikMamba achieves remarkable recognition performance, surpassing the previous state-of-the-art by 1.45\%, 7.22\%, 0.15\%, and 3.92\% on the PAF, HARDVS, DVS128, and E-FAction datasets, respectively. The code is available at \url{https://github.com/Typistchen/SpikMamba}.

\end{abstract}




\maketitle

\begin{figure}
    \centering
    \includegraphics[scale=0.43]{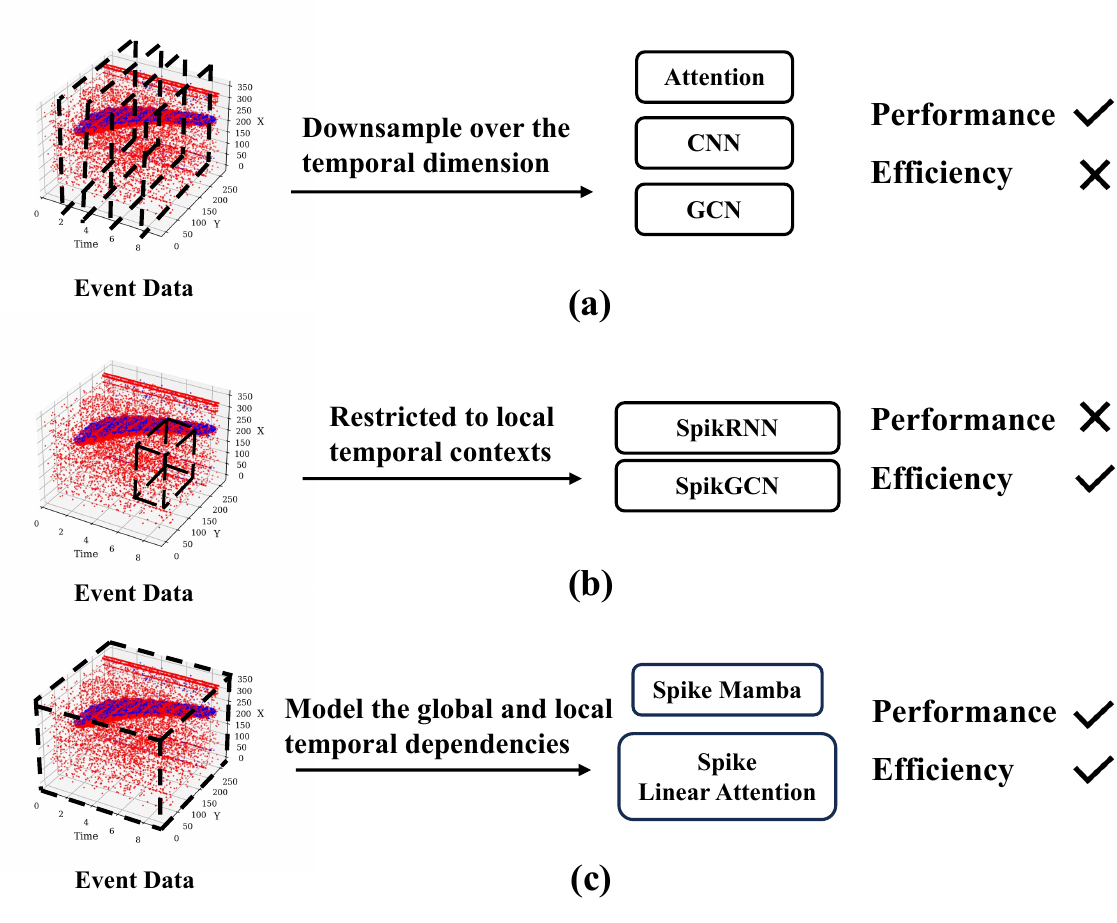}\\
    \caption{ 
    Overview of (a) ANN methods, (b) SNN methods, and (c) ours for event-based HAR. (a) ANN methods downsample the event data temporally to reduce the heavy computation, and use an attention mechanism, convolutional neural network (CNN), and graph neural network (GCN) to extract features from spatially sparse events, which achieves high performance. However, fine-grained information about human actions, which could improve the model performance, is lost. (b) SNN methods use SpikRNN and SpikGCN to effectively extract features from spatially sparse events, however, the computations are usually restricted to local temporal contexts, which leads to a loss of global temporal dependency for accurately recognizing human actions.  (c) We combine Mamba and window-based linear attention into SNN to efficiently model the global and local temporal dependencies of the event data, and accurately recognize human actions.}
    \label{fig:intro}
    \vspace{-0.2cm}
\end{figure}

\section{Introduction}

Human action recognition (HAR) aims to classify human activities and movements \cite{kong2022human,liu2024lightweight}. It has been applied to various domains such as robot navigation \cite{mavrogiannis2023core, sun2022human}, healthcare \cite{carreira2017quo}, and abnormal human behaviour recognition \cite{lentzas2020non, pareek2021survey}. Most HAR methods are developed for RGB images. Although they achieve high performance, human privacy information is inevitably recorded, \textit{e.g.}, facial features, which presents challenges and concerns for deploying under privacy-sensitive environments \cite{rajpoot2017video, slobogin2002public}. Therefore, we pose a question: can we design a framework that effectively protects user privacy and accurately recognizes human actions?


Event cameras are novel sensors inspired by the working mechanism of the human retina \cite{amir2017low, gao2023action, liu2021event}. Unlike traditional RGB cameras that record all pixel intensities, event cameras asynchronously and sparsely detect changes in light intensity with microsecond-level temporal resolution and a high dynamic range \cite{pan2019bringing,9252186,yang2023event,ECDDP}. This means that privacy-related features are usually discarded, \textit{e.g.}, facial textures. While using event camera data for human action recognition (HAR) can address user privacy concerns, it introduces new challenges for HAR frameworks: 
1) The event stream is spatially sparse, requiring the model to associate events from different times to capture meaningful features.
2) The event stream has a high temporal resolution, resulting in an excessive number of events that require efficient processing. In this paper, we aim to design an event-based HAR framework that addresses these challenges for HAR with high performance.

Existing event-based HAR methods are developed based on artificial neural networks (ANN) \cite{gao2023action, pradhan2019n,xie2022vmv, calabrese2019dhp19} or spiking neural networks (SNN) \cite{cao2015spiking, hunsberger2015spiking, bu2023optimal, meng2022training, wang2022signed,fang2021incorporating}. To handle the spatial sparsity of event camera data, ANN-based methods (Fig. \ref{fig:intro}(a)) often use attention mechanisms, convolutional neural network, or graph convolutional network to enhance feature extraction from the sparse event data. For computational efficiency, these methods downsample the event data over the temporal dimension, \textit{e.g.}  \cite{calabrese2019dhp19} uses event data of 48ms duration for every 0.35 seconds. However, the event data downsampling loses fine-grained information about human actions, which could enhance model performance. 

By design, SNNs (Fig. \ref{fig:intro}(b)) effectively handle the spatial sparsity of event camera data through event-driven computation on temporal dynamics, integrating event features over time to form a coherent understanding of the scene. However, the computations of existing SNN-based methods are usually restricted to local temporal contexts for computational efficiency, and they lose the global temporal dependency of event data for accurately recognizing human actions.
While methods such as attention mechanisms can be applied to dynamically capture global temporal dependencies, doing so often reduces the efficiency of SNNs.

Luckily, recent advancements in state space models \cite{gu2021efficiently, smith2022simplified, fu2022hungry, mehta2022long}, such as Mamba \cite{gu2023mamba}, suggests an efficient solution for dynamically modeling data with a high temporal resolution, offering an alternative to attention mechanisms. Motivated by the success of Mamba and SNNs, we propose combining these approaches to efficiently and accurately recognize human actions using event data. To this end, we are the first to introduce the SpikMamba which has the two key designs for event-based HAR. 

To address the spatial sparsity and high temporal resolution of event camera data, we model the global and local temporal dependencies of the event data (Fig. \ref{fig:intro}(c)). First, we construct a Mamba block in spike form (only 0s and 1s) to globally model the interdependencies among event data. Second, to enhance the locality of spike features, we apply a spike-based linear attention mechanism to the event data across different temporal windows. To validate the effectiveness of our framework, we experiment with common event-based HAR datasets, demonstrating that our method surpasses previous state-of-the-art approaches.

In summary, our main contributions are: 
\begin{itemize}
    \item We propose a SpikMamba framework to effectively and accurately recognize human actions using event data.
    \item We explore Mamba and window-based linear attention spike-based mechanisms for modeling global and local temporal dependencies of the event data.
    \item We experiment with common event-based HAR data to demonstrate our superior performance compared to existing state-of-the-art algorithms.
\end{itemize}

\begin{figure*}
    \centering
    \includegraphics[scale=0.35]{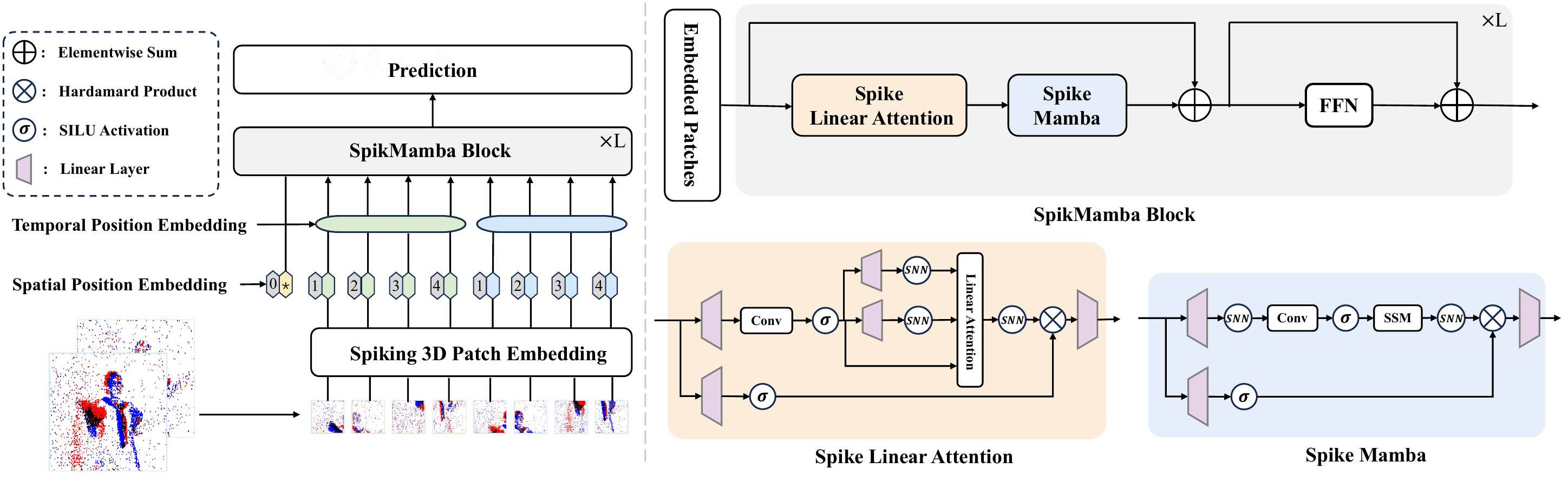}\\
    \caption{ The overview of SpikMamba. 
    We represent event data as three channel event images, and predict the action class of the event data with two modules:  1) Spiking 3D patch embedding. It divides event frames into patches and project the patches to spike-form features.
    2) SpikMamba blocks. It consists of a window-based spike linear attention layer and a spike Mamba layer to model the local and global temporal dependencies of the event data. We show the architecture of the SpikMamba block at the top right, and the architecture of the spike linear attention layer and the spike Mamba layer in the bottom right.
    Finally, through the prediction layer, the  embedding of the last  SpikMamba block is pooled and projected to the action class.
    }
    \label{fig:framework1}
    \end{figure*}
\section{Related Work}

In this section, we briefly introduce ANN for Event-based HAR, SNN for Event-based HAR, and the state space model.

\noindent{\bf ANN for Event-based HAR.} ANN-based methods typically use CNNs \cite{gao2023action, pradhan2019n}, ViTs \cite{sabater2022event, xie2023event}, and GCNs \cite{xie2022vmv, calabrese2019dhp19} to extract sparse event data feature. EV-ACT \cite{gao2023action} employs a CNN with spatial-temporal attention for action recognition, while \cite{pradhan2019n} adapts event data to CNNs using event memory surfaces. ViTs \cite{xie2022vmv, calabrese2019dhp19} employ patch-based and voxel transformer encoders for efficient spatio-temporal feature extraction, and GCNs  \cite{ xie2023event} manage the sparse, asynchronous structure. However, most ANN-based HAR methods overlook spatial sparsity and high temporal resolution. Our SpikMamba network tackles both issues effectively.


\noindent{\bf SNN for Event-based HAR.} SNNs for Event-based HAR. Spiking Neural Networks (SNNs) \cite{cao2015spiking, hunsberger2015spiking, bu2023optimal, meng2022training, wang2022signed,fang2021incorporating} differ from traditional deep learning models, which use continuous decimal values, while SNNs utilize discrete spike sequences. This makes SNNs well-suited for processing temporal data, leading to their use in event-based HAR \cite{amir2017low, liu2021event, george2020reservoir,lee2018training,soures2019deep}. However, SNN-based HAR methods often lose fine-grained action details due to event data downsampling. In contrast, our SpikMamba effectively combines Mamba and window-based linear attention mechanisms in spike form to model global and local temporal dependencies.


\noindent{\bf State Space Model.} The state-space model \cite{gu2021efficiently} (S4) serves as an alternative to CNNs and Transformers for long-range dependency modeling. Mamba has been applied to event data \cite{qin2024mamba,wang2024mambaevt}, with \cite{qin2024mamba} integrating a spiking front-end for temporal processing and \cite{wang2024mambaevt} using a linear complexity state-space model for tracking. Our research combines Mamba’s time-series strengths with SNNs’ efficiency in sparse event data, proposing the SpikMamba network.

\section{Method}

\subsection{Preliminaries}

\textbf{SNN.} The core of SNN is spiking neurons that receive the input $X[t]$ and accumulate membrane potentials $H[t]$ at each time step $t$. When the membrane potential $X[t]$  exceeds a threshold $V_{\text{th}}$, a spike $S[t]$ is triggered, and the membrane potential is reset. In our work, we employ the Leaky Integrate-and-Fire (LIF) spiking neurons \cite{wu2018spatio}. The mathematical representations of LIF are summarized as follows:
\begin{align}
    H[t] &= V[t-1] + \frac{1}{\tau}(X[t] - (V[t-1] - V_{\text{reset}}) \ , \label{lif_(1)} \\
    S[t] &= \Theta(H[t] - V_{\text{th}})  \label{lif_(2)} \ , \\
    V[t] &= H[t] (1 - S[t]) + V_{\text{reset}}S[t]  \ , \label{lif_(3)}
\end{align}
where $\tau$ is the membrane time constant, $\Theta(\cdot)$ is the Heaviside step function that fires a spike (outputs 1) if $H[t] - V_{\text{th}} \geq 0$, and $V[t]$ is the membrane potential which resets to  $V_{\text{reset}}$ if a spike is fired.

\noindent \textbf{Mamba.} Mamba is inspired by continuous system that maps a sequence $X$ to $Y$ using a hidden time state $Z$. At each time step $t$, the mapping is calculated using the state-space equations:
\begin{align}
Z'[t] &= AZ[t] + BX[t] \ , \\
Y[t] &= CZ[t] \ ,
\end{align}
where $Z'[t]$ is a placeholder variable, $A$ is the system evolution matrix, and $B$ and $C$ are the projection matrices. In Mamba, the state-space equations are discretized, by using the zero-order hold (ZOH) method to transform the parameters continuous system $A$ and $B$ into their discrete versions $\overline{A}$ and $\overline{B}$ with a timescale parameter $\Delta$ to control the step size of the discretization process,
\begin{align}
    \overline{A} &= \text{exp}(\Delta A) \ , \\
    \overline{B} &= (\Delta A)^{-1} (\text{exp}(\Delta A) - \textbf{I}) \cdot \Delta B \ ,
\end{align}
where $I$ is an identity matrix, and $\cdot$ represents elementwise multiplication. Mamba also makes the parameters $B$, $C$, and $\Delta$ dependent on input $X[t]$ to calculate $\overline{A}$ and $\overline{B}$. Additionally, it uses a global convolution shared across different time steps to compute the output $Y[t]$. We refer the reader to \cite{gu2023mamba} for more details.








\subsection{SpikMamba}
We use the representation from \cite{zhou2024exact} that transforms the event data into three channel event images $X \in \mathbb{R}^{3 \times T \times H \times W}$, where $T$, $H$, and $W$  are the temporal dimension, height, and width of the event images. We predict the action class of the event images $X$ with our SpikMamba (Fig. \ref{fig:framework1}), which has two main modules:
i) Spiking 3D patch embedding. It splits event frames $X$ into patches to calculate patch embeddings $P$ with SNN.
ii) SpikMamba block. It encapsulates window-based linear attention and Mamba into SNN to model local and global temporal dependency of event data for HAR from the patch embeddings. Last, the embeddings produced by the SpikMamba blocks are pooled and then projected to the action class with the use of a final linear layer for classification.

\noindent \textbf{Spiking 3D Patch Embedding.} As shown in Fig. \ref{fig:framework1}, we first divide the event frames $X$ into patches and then project them into spike-form features. Similar to ViT, we use a convolution layer with shared parameters across patches to calculate the patch embedding $P$,
\begin{align}
    P = \text{SL}_{\text{patch}}\big(\text{BN}(\text{Conv3d}(X))\big) + \text{PE} \ ,
\end{align}
where $\text{SL}_{\text{patch}}(\cdot)$, $\text{BN}(\cdot)$, $\text{Conv3d}(\cdot)$ are the spike layer, batch normalization layer, and convolution 3D layer with a stride of 1 $\times$ 8 $\times$ 8 and kernel size of 1 $\times$ 8 $\times$ 8, and $\text{PE}$ is the positional embedding that introduces inductive bias on the spatial and temporal dimensions to the patch embeddings. Though more than one 3D convolution layer can be used to progressively compute the patch embedding, our experiments show that a single Conv3D layer is sufficient to accurately recognize human actions and is the most efficient option.

\noindent \textbf{SpikMamba Block.} The patch embeddings $P$ of event frames $X$ are sent to $N$ SpikMamba blocks. In a SpikMamba block, it includes a window-based spike linear attention layer $\text{SpikeSLA}(\cdot)$, a spike Mamba layer $\text{SpikMamba}(\cdot)$, and a feedforward network $\text{FFN}(\cdot)$. For simplicity, we describe the calculations of a single Mamba block without differentiating between the Mamba block indices:
\begin{align}
    P_\text{local} &= \text{SpikeSLA}(P) \ , \\
    P_\text{global} &= \text{SpikMamba}(P_\text{local}) + P_\text{local} \ , \\
    P_\text{out} &= \text{FFN}(P_\text{global}) + P_\text{global} \ , 
\end{align}
where $P_\text{local}$, $P_\text{global}$, and $P_\text{out}$ are the output patch embeddings generated from the respective layers.

In the window-based spike linear attention layer, we reshape the patch embedding $P$ to divide it into different windows, and project the patch embedding into spike form query $Q$, key $K$, and value $V$ using linear layers. We use spike-form query and key, while the continuous value is used to improve feature representation,  
\begin{align}
Q &= \text{SL}_{\text{q}} \big(\text{Linear}_{\text{q}}(\text{Reshape}_\text{window}(P)) \big) \ , \\
K &= \text{SL}_{\text{k}} \big(\text{Linear}_{\text{k}}(\text{Reshape}_\text{window}(P)) \big) \ , \\
V &=  \text{Linear}_{\text{v}}(\text{Reshape}_\text{window}(P)) \ ,
\end{align}
where $\text{SL}_{\text{q}}(\cdot)$ and $\text{SL}_{\text{k}}(\cdot)$ are spike layers responsible for processing query and key, $\text{Linear}_{\text{q}}(\cdot)$, $\text{Linear}_{\text{k}}(\cdot)$, and $\text{Linear}_{\text{v}}(\cdot)$ are linear layers, and $\text{Reshape}\text{window}(\cdot)$ is the window reshape layer. Then, we calculate the embedding $P_\text{att}$ using a linear attention layer $\text{LinearAtt}(\cdot,\cdot,\cdot)$ from \cite{han2024demystify} and a spike layer $\text{SL}_\text{att}(\cdot)$,
\begin{align}
    P_\text{att} = \text{SL}_{\text{att}}(\text{LinearAtt}(Q, K, V)) \ .
\end{align}
The $P_\text{att}$ is projected to the patch embedding $P_\text{local}$ with a linear output layer $\text{Linear}_\text{out}
(\cdot)$, reshaped back with $\text{Reverse}_\text{window}(\cdot)$, and undergoes a Hadamard product with $P$, 
\begin{align}
    P_\text{local} = \text{Reverse}_\text{window}(\text{Linear}_\text{out}(P_\text{att}))  \circ P\ .
\end{align}

To model the temporal global dependency in $P_\text{local}$, our spike Mamba layer $\text{SpikMamba}(\cdot)$ uses a linear layer $\text{Linear}_\text{m}(\cdot)$ and a 1D convolution layer $\text{Conv1D}_\text{m}(\cdot)$ with spike layers $\text{SL}_\text{m1}(\cdot)$ and $\text{SL}_\text{m2}(\cdot)$
to expand the dimension of $P_\text{local}$,
\begin{align}
    P_\text{global}' = \text{SL}_\text{m2}(\text{Conv1D}_\text{m}(\text{SL}_\text{m1}(\text{Linear}_\text{m}(P_{\text{local}})))) \ ,
\end{align}
and predict the evolution matrices $A$, $B$, and the timescale parameter $\Delta$ of the state-space equations in Mamba with $\text{Linear}_\text{B}(\cdot)$, $\text{Linear}_\text{C}(\cdot)$, and $\text{Linear}_{\Delta}(\cdot)$,
\begin{align}
    B &= \text{Linear}_\text{A}(P_\text{global}') \ , \\
    C &= \text{Linear}_\text{B}(P_\text{global}') \ ,\\
    \Delta &= \text{log}(1 + \text{exp}(\text{Linear}_\text{A}(P_\text{global}') + \text{bias}_{\Delta}) \ ,
\end{align}
where $\text{bias}_{\Delta}$ is a trainable bias. With trainable system evolution parameter $A$, $\Delta$ and $B$, $A$ and $B$ are discretized into their respective forms $\overline{A}$ and $\overline{B}$ \cite{gu2023mamba}, the state space equation $\text{SSM}(\cdot, \cdot, \cdot, \cdot)$ is calculated with a spike layer $\text{SL}_{\text{ssm}}(\cdot)$, and undergoes a Hadamard product with $P_\text{local}$,
\begin{align}
    P_\text{global} = \text{SL}_{\text{ssm}}(\text{SSM}(\overline{A},\overline{B}, C, P_\text{global}')) \circ P_\text{local} \ ,
\end{align}
and the output $P_\text{global}$ is then sent to the feedforward layer $\text{FFN}(\cdot)$.

\noindent \textbf{Prediction.} We pool the patch embeddings $P_\text{out}$ from the SpikMamba block with a global average pooling $\text{GAP}(\cdot)$, and predict the human action $y$ using a linear layer $\text{Linear}_{\text{predict}}(\cdot)$,
\begin{align}
    y = \text{Linear}_{\text{predict}} (\text{GAP}(P_\text{out})) \ .
\end{align}
The prediction $y$ is optimized with the ground truth human action class and cross-entropy loss during training.

\vspace{-0.2cm}
\section{Experiments}

\subsection{Dataset and Implementation details}

\begin{figure}
    \centering
    \includegraphics[scale=0.7]{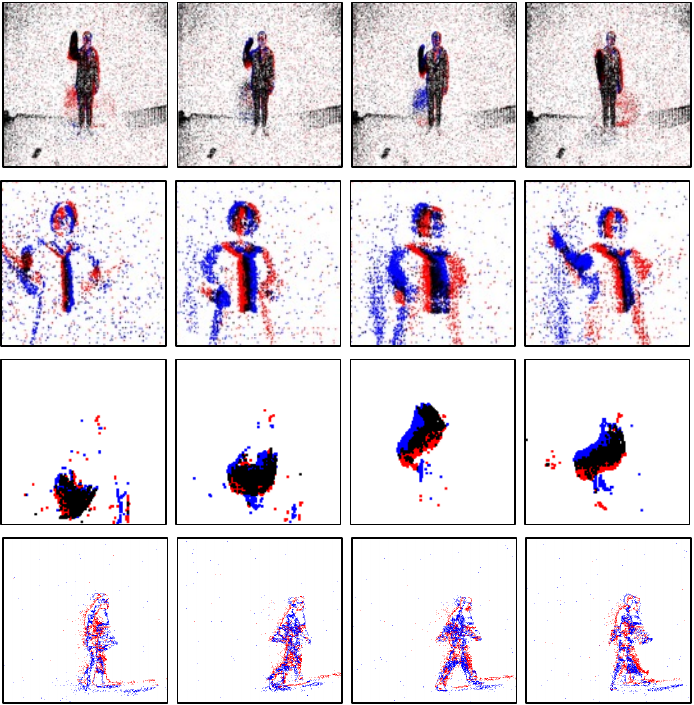}\\
    \caption{
    Examples of event data. From the first to last rows, they are event frames of `throw' in the PAF dataset, `type clamp back' in the HARDVS dataset, `right arm clockwise' in the DVSGesture dataset and `slow walking' in the E-FAction dataset.
   }\label{fig:protect}
    
\end{figure}

\noindent\textbf{Dataset.} We use four datasets to evaluate the performance of our model, SpikMamba. These datasets include PAF \cite{miao2019neuromorphic}, HARDVS \cite{wang2024hardvs}, DVSGesture \cite{amir2017low}, and E-FAction \cite{zonglin2024event}. Specifically,
1) PAF \cite{miao2019neuromorphic} is a human action dataset collected using the DVSIS346 event camera, containing 10 categories of actions with 45 samples per category.
2) HARDVS \cite{wang2024hardvs}, a recently released dataset, boasts the largest number of action categories and samples, totaling 300 categories and 107,646 recordings. 
3) DVSGesture \cite{amir2017low} captures hand and arm movements, containing 11 action categories with a resolution of 128$\times$128. 
4) E-FAction \cite{zonglin2024event} dataset has 128 human action classes,  totaling 1024 recordings with a resolution of 346$\times$260. 
We show examples of the datasets in Fig. \ref{fig:protect}.

\begin{table}[]
\caption{
Comparison with state-of-the-art models for event-based action recognition on PAF, HARDVS, DVSGesture, and E-FAction datasets. The models are evaluated with accuracy (ACC), and we show the type of model, \textit{i.e.,} ANN or SNN. The method with the highest accuracy is in bold.
}
\label{tab:1_compration}
\setlength{\tabcolsep}{7.91pt}
\begin{tabular}{cccc}
\toprule
\multicolumn{1}{c}{Dataset} & \multicolumn{1}{c}{Model} & \multicolumn{1}{c}{SNN}   &      \multicolumn{1}{c}{Acc(\%)}                 \\ 
\midrule

\multirow{8}{*}{PAF}            & HMAX SNN \cite{xiao2019event}              &   \CheckmarkBold   & 55.00                    \\
                                & STCA  \cite{gu2019stca}                  &  \CheckmarkBold    & 71.20                       \\
                                & Motion SNN \cite{liu2021event}             &  \CheckmarkBold    & 78.10                          \\
                                & MST    \cite{wang2023masked}                &   \CheckmarkBold   & 88.21                         \\
                                & Swin-T (BN)    \cite{wang2023masked}         &  \CheckmarkBold   & 90.14                         \\
                                & EV-ACT      \cite{gao2023action}             &  \XSolidBrush             & 92.60                         \\
                                & ExACT       \cite{zhou2024exact}            &    \XSolidBrush                & 94.83                        \\
                                & \textbf{SpikMamba(Ours)}&    \CheckmarkBold             & \textbf{96.28}                  \\ 
                                \midrule
                                
\multirow{12}{*}{HARDVS}        & X3D    \cite{feichtenhofer2020x3d}                & \XSolidBrush    & 45.82                   \\
                                & SlowFast  \cite{feichtenhofer2019slowfast}             & \XSolidBrush    & 46.54                       \\
                                & ACTION-Net \cite{wang2021action}             &  \XSolidBrush   & 46.85                    \\
                                & R2Plus1D   \cite{tran2018closer}            & \XSolidBrush    & 49.06                      \\
                                & ResNet18    \cite{he2016deep}           &  \XSolidBrush   & 49.20                    \\
                                & TAM      \cite{liu2021tam}              & \XSolidBrush    & 50.41                       \\
                                & C3D  \cite{tran2015learning}                  &  \XSolidBrush   & 50.52                       \\
                                & ESTF    \cite{wang2024hardvs}               & \XSolidBrush    & 51.22                         \\
                                & Video-SwinTrans \cite{liu2022video}       & \XSolidBrush    & 51.91                         \\
                                & TSM     \cite{lin2019tsm}               & \XSolidBrush    & 52.63                        \\
                                & ExACT      \cite{zhou2024exact}            & \XSolidBrush    & 90.10                        \\
                                & \textbf{SpikMamba(Ours)}  & \CheckmarkBold  & \textbf{97.32}       \\ 
                                \midrule
\multirow{9}{*}{DVSGesture} & Time-surfaces    \cite{maro2020event}&   \CheckmarkBold & 90.62                          \\
                                & SNN eRBP    \cite{kaiser2019embodied}         &  \CheckmarkBold    & 92.70                          \\
                                & Slayer  \cite{shrestha2018slayer}             &  \CheckmarkBold     & 93.64                          \\
                                & DECOLLE  \cite{kaiser2020synaptic}            &  \CheckmarkBold    & 95.54                            \\
                                & EvT \cite{sabater2022event}                 &  \CheckmarkBold   & 96.20                           \\
                                & TBR       \cite{innocenti2021temporal}           &  \XSolidBrush     & 97.73                         \\
                                & EventTransAct   \cite{de2023eventtransact}     &  \XSolidBrush    & 97.92                          \\
                                & ExACT   \cite{zhou2024exact}             &  \XSolidBrush    & 98.86                        \\
                                & \textbf{SpikMamba(Ours)}& \CheckmarkBold  & \textbf{99.01}              \\ 
                                \midrule

\multirow{6}{*}{E-FAction} 
                                & CLIP-L    \cite{radford2021learning}          & \XSolidBrush   & 61.90                          \\
                                & ResNet3D-N  \cite{kim2021n}        &  \XSolidBrush       & 65.60                         \\
                                & ResNet3D-K  \cite{kay2017kinetics}        &  \XSolidBrush      & 66.30                             \\
                                & MASTAF    \cite{liu2023mastaf}          &  \XSolidBrush      & 67.10                          \\
                                 & ExACT    \cite{zhou2024exact}          &  \XSolidBrush      & 67.93                          \\
                                & \textbf{SpikMamba(Ours)} & \CheckmarkBold  & \textbf{71.02}                        \\ 
                                \bottomrule
                                
\end{tabular}

\end{table}
\noindent\textbf{Implementation details.} We employ the AFE representation \cite{zhou2024exact} to compress the event stream into event frames. 
Our SpikMamba model has a layer of spiking 3D patch embedding and two layers of SpikMamba blocks for feature extraction. We use a hidden state dimension of 256, and expand the state dimension for state space equations to 256 with the $\text{Linear}_{\text{m}}(\cdot)$. The state-space equation operates in a dimension of 2048. The hidden dimension of the feedforward networks is 1024. In training, we use the Adam optimizer with a weight decay of $2e^{-4}$. The learning rate is initialized to $1e^{-5}$, and we adopt the CosineAnnealingLR, \cite{loshchilov2016sgdr} with a minimum learning rate of $1e^{-6}$. Our model was trained on two NVIDIA 4090 GPUs for 100 epochs with batch size 32. Our code will be made available online for future studies and comparisons.

\subsection{Comparison with SOTA Methods} 
Tab. \ref{tab:1_compration} shows the performance of our proposed SpikMamba on the PAF, HARDVS, DVSGesture, and E-FAction datasets for event-based action recognition tasks. We compare it with state-of-the-art methods on all these datasets. The method with the highest accuracy is highlighted in bold in the table.


Our findings are as follows: 1) Our method has the best accuracy of 96.28\%, 97.32\%, 99.01\%, and 71.02\%  across the four datasets. 2) Compared to ANN-based ExACT, which has the second highest accuracy, our method shows accuracy improvements of 1.45\%, 7.22\%, 0.33\%, and 3.09\%. 3) On the HARDVS dataset, which has the largest number of complex and diverse human actions, our SpikMamba and ExACT show significant improvements over other methods, showing accuracy increases of more than 35\%. Additionally, our SpikMamba further improves ExACT's accuracy by 7.22\%. 4) On the DVSGesture dataset, the highest accuracy of the state-of-the-art method is already 98.86\%, but our SpikMamba increases it to 99.01\%. 5) Compared to the SNNs with the second highest accuracy across the four datasets, our SpikMamba improves the accuracy  by 6.14\% and 2.81\% on PAF and DVSGesture dataset, which is the first SNN method that is better than the ANN method.


\begin{figure*}
    \centering
    \includegraphics[scale=0.54]{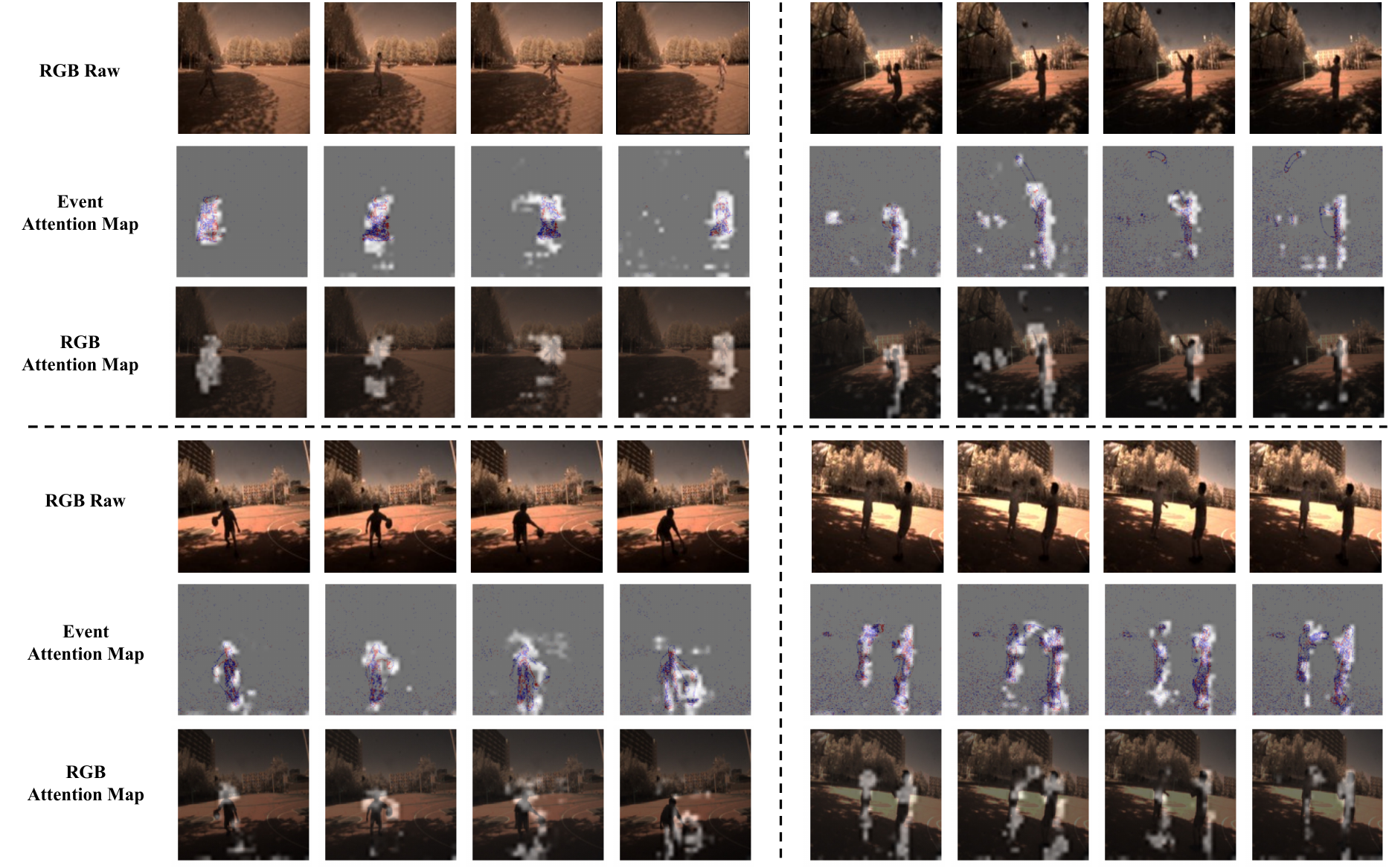}\\
    \caption{ Attention map examples of SpikMamba.  High-attention regions are marked in white, and low-attention regions are marked in black. The attention map indicates that our SpikMamba focuses on image regions with human action effectively. Please view in colour on the screen.
   }\label{fig:attention mamp}
    
\end{figure*}

\subsection{Ablation Studies}

\begin{table}[]
\caption{Ablation study of SpikeSLA and SpikMamba layers.  The highest accuracy is in bold.}
\label{tab:2_ablation}
\setlength{\tabcolsep}{2.74pt}
\begin{tabular}{cccccc}
\toprule
SpikeSLA &    SpikMamba        & HARDVS & PAF   & DVSGesture & E-FAction \\ 
\midrule
\CheckmarkBold & \XSolidBrush    & 97.12  & 95.33 & 98.17          & 70.66     \\
\XSolidBrush  &  \CheckmarkBold  & 77.85  & 71.44 & 80.56          &  54.85    \\ 
\CheckmarkBold & \CheckmarkBold              & \textbf{97.32}  & \textbf{96.28} & \textbf{99.01}          & \textbf{71.02}     \\ 
\bottomrule
\end{tabular}

\end{table}

We ablate SpikeSLA and SpikMamba layers in Tab. \ref{tab:2_ablation}. We found: 1) When using only the SpikeSLA layers of our model, the network significantly loses the ability to capture long-term/global information from high temporal resolution event data, and has 97.12\%, 95.33\%, 98.17\%, and 70.66\% accuracy. 2) When removing the SpikeSLA layers from our model, we observe a significant accuracy drop. The average decrease is 19.73\%. Given that the action durations recorded in the four datasets primarily range from 5 to 7 seconds, key frames of the action are likely short-term that   constitute the main features of the action. Consequently, when the SpikeSLA is removed from our model, the network cannot effectively enhance the feature locality for HAR. 3) The model with SpikeSLA and SpikMamba layers  efficiently and accurately models the global and local temporal dependencies of the event data, and has the best performance. 

\subsection{Discussion}

\noindent \textbf{Attention Map.} 
In Fig. \ref{fig:attention mamp}, we illustrate attention maps from the final SpikMamba block at the last time step.  For clarity, we provide attention maps on RGB images generated by SpikMamba. High attention regions are marked in white, while low attention regions are marked in black. Our SpikMamba effectively captures image regions with human actions.


\begin{table}[]
\caption{Computational Efficiency. We compare SpikMamba with ExACT and EvT, which are state-of-the-art ANN and SNN methods.
}
\label{tab:2_Efficiency}
\setlength{\tabcolsep}{13pt}

\begin{tabular}{ccc}
\hline
Model     & GLOPs & \#Params. \\ \hline
ExACT \cite{zhou2024exact} & 1.1   & 2.13M     \\
EVT  \cite{sabater2022event}     & 0.2   & 0.48M     \\
\textbf{SpikMamba(Ours)} & \textbf{0.12}  & \textbf{0.18M}     \\ \hline
\end{tabular}

\end{table}

\noindent \textbf{Computational Efficiency.}
We compare SpikMamba with the state-of-the-art ANN and SNN methods that are ExACT and EvT on computational efficiency in Tab. \ref{tab:2_Efficiency}. Our method has 0.18M parameters, which is 1.95M and 0.30M less than  ExACT and EvT. The FLOPs of SpikMamba, ExACT, and EvT are 0.12, 1.1, and 0.2 GFLOPs. Our SpikMamba combines SNN and Mamba to efficiently capture global dependencies in event data and uses a spiking window-based linear attention mechanism to model the event data local dependency, striking a balance between computational efficiency and performance in HAR. Our method has the fewest parameters and FLOPs, while also achieving better HAR performance than the best state-of-the-art ANN and SNN methods.

\noindent \textbf{ANN and SNN.} 
To explore the performance of SpikMamba, we removed the SNN layers module. It creates an ANN model based on window-based linear attention and Mamba. The results for four datasets are 94.53\%, 92.47\%, 98.01\%, and 67.77\%. Compared to SpikMamba, the ANN method shows a decrease in performance across all four datasets, and the average is 2.71\%. It is evident that SNN-based Mamba and linear attention are more suitable for event data. We believe this is because of the alignment between the sparsity of SNNs and the sparsity of event data, enabling SNN-based Mamba and linear attention to effectively and accurately model the global and local dependencies of the event data for HAR.

\section{Conclusion}

In this paper, we propose SpikMamba for event data-based Human Activity Recognition (HAR). Using event data for HAR presents challenges in effectively capturing meaningful features from spatially sparse and high temporal resolution event data. By leveraging the energy efficiency of Spiking Neural Networks (SNN) and the long sequence modeling capabilities of Mamba, SpikMamba effectively captures global dependencies from sparse and high temporal resolution event streams. Additionally, a spiking window-based linear attention mechanism is proposed to enhance the locality of event data modeling for HAR. Experiments on common event-based HAR datasets demonstrate our superior performance compared to existing state-of-the-art ANN and SNN methods.

\section{Acknowledgments}
This work was supported in part by the Beijing Institute of Technology (BIT) Research Fund Program for Young Scholars, the BIT Special-Zone, and National Natural Science Foundation (NSFC) of China under grants 62302045 and 62202087.


\bibliographystyle{ACM-Reference-Format}
\bibliography{samplebase.bib}

\end{document}